\documentclass[letterpaper, 10 pt, conference]{ieeeconf}  % Comment this line out if you need a4paper

\IEEEoverridecommandlockouts                              % This command is only needed if 
\usepackage{array}
\usepackage{booktabs}
\usepackage{graphicx}
\usepackage{color}
\usepackage[ruled,vlined]{algorithm2e}
\usepackage{cite}
\usepackage{algorithmic}
\usepackage{subcaption}
\usepackage{multirow}
\usepackage{amsmath,amssymb,amsfonts}
\usepackage{arydshln}
\usepackage{diagbox}
\usepackage{graphicx}
\usepackage{footmisc}
\usepackage[bookmarks=false]{hyperref}
\usepackage{float}
\usepackage{authblk}
\usepackage{placeins}

\title{\LARGE \bf
Deep Inertial Odometry with Accurate IMU Preintegration
}

\author{Rooholla Khorrambakht$^{1}$ , Chris Xiaoxuan Lu$^{2}$, Hamed Damirchi$^{1}$, Zhenghua Chen$^{3}$, Zhengguo Li$^{3}$ %
\thanks{$^{1}$Rooholla Khorrambakht and Hamed Damirchi are with Faculty of Electrical Engineering and Computer Science, K.~N.~Toosi University of Technology, lran,
        {\tt\small r.khorrambakht, hdamirchi@email.kntu.ac.ir}}%
\thanks{$^{2}$Chris Xiaoxuan Lu is with the School of Informatics, University of Edinburgh, United Kingdom,
    	{\tt\small xiaoxuan.lu@ed.ac.uk}}
\thanks{$^{3}$Zhenghua Chen and Zhengguo Li are with the Institute for Infocomm Research (I2R), A*STAR, Singapore,
    	{\tt\small Chen\_Zhenghua, ezgli@i2r.a-star.edu.sg}}
    	}
\begin{document}

\maketitle

\begin{abstract}
Inertial Measurement Units (IMUs) are interceptive modalities that provide ego-motion measurements independent of the environmental factors. They are widely adopted in various autonomous systems. Motivated by the limitations in processing the noisy measurements from these sensors using their mathematical models, researchers have recently proposed various deep learning architectures to estimate inertial odometry in an end-to-end manner. Nevertheless, the high-frequency and redundant measurements from IMUs lead to long raw sequences to be processed. In this study, we aim to investigate the efficacy of accurate preintegration as a more realistic solution to the IMU motion model for deep inertial odometry (DIO) and the resultant DIO is a fusion of model-driven and data-driven approaches. The accurate IMU preintegration has the potential to outperform numerical approximation of the continuous IMU model used in the existing DIOs. Experimental results validate the proposed DIO.
\end{abstract}
% \begin{IEEEkeywords}
% 	Inertial Odometry, IMU Pretintegrated Factors, Artificial Intelligence, Multimodal Fusion.
% \end{IEEEkeywords}

\IEEEpeerreviewmaketitle

\section{Introduction}
\label{sec:introduction}
Inertial Odometry (IO) is concerned with the estimation of ego-pose transformations using raw IMU measurements. Performing IO based on the mathematical model of an IMU leads to large accumulative errors in long runs and is only reliable for short time-intervals between the measurements from other complementary sensors in a fusion setup. However, learning the inertial odometry encapsulates the IO problem in a sequence modeling setup and is motivated by the flexibility of deep learning in exploiting the complex motion patterns in the data as pseudo measurements for keeping the error in check. 

% \begin{figure}
%      \centering
%      \begin{subfigure}[b]{0.45\textwidth}
%         %  \centering
%          \includegraphics[width=1\textwidth]{data/intro-image.pdf}
%         %  \caption{Seq. 10}
%      \end{subfigure}
%         \caption{Accurate preintegration performs suitably for both fast and slow motion domains.}
%         \label{fig:intro-image}
% \end{figure}
IMU measurements are produced at high frequencies, which leads to long sequences for relatively short time periods. Processing these long sequences using Recurrent Neural Networks (RNNs) is challenging and is prone to washout \cite{zhao2020rnn,trinh2018learning}, while processing them using Convolutional Neural Networks (CNNs) requires deep architectures to cover large enough receptive fields. Although Wavenets have been employed to address such problems \cite{chen2020deep}, they have not yet found widespread adoption due to their specialized architecture. As another alternative to this solution, we proposed exploiting preintegration as a model-aware preprocessing step for compressing the temporal dimension of the raw motion measurements from the IMU \cite{khorrambakht2020preintegrated}. As pointed out in the literature \cite{kaufmann2020deep}, correct inductive biases, and proper intermediate representations can lead to significant performance boosts. By introducing preintegration as a method of extracting intermediate representations, we achieved significant performance improvements while also reducing the computational load. Furthermore, the produced intermediate representations do not impose any architectural constraints on the deep learning models. 

However, the validity of this compression relies on the correctness of the assumptions made for formulating the motion preintegration. The formulation presented in \cite{khorrambakht2020preintegrated} relied on the derivations of \cite{forster2017onmanifold} which is based on the assumption that the angular velocity in body frame and linear acceleration in world frame between two consecutive samples remain constant. This assumption holds when the motion is not highly dynamic or when the IMU sample rate is high. The authors of \cite{henawy2019accurate} addressed this issue by exploiting the linear switched systems theory and proposed an accurate preintegration formulation for Visual Inertial Odometry (VIO) applications. This study aims to adopt the presented formulation by \cite{henawy2019accurate} in computing the PreIntegrated (PI) features in our deep learning setup and study its effectiveness in two dynamic and moderate motion domains. As such, the model-driven and data-driven approaches co-exist under one roof. Using real-world datasets, we show that the two features lead to similar performances in moderate motions. However, at higher speeds and dynamic motions, accurate preintegration yields better performance. Nevertheless, the fusion of IMU preintegration and deep learning results in a deep inertial odometry (DIO) that is useful for the emerging cognitive navigation for robots. The cognitive navigation has the potential to replace the popular metric navigation due to the development of artificial intelligence \cite{wolbers2014challenges}.   

The remainder of this report is structured as follows. Section \ref{sec:preintegration} presents the preintegration theory and the formulation of its accurate solution. Next, the experimental setups, datasets, and results are introduced in Section \ref{sec:experiments} and the concluding remarks are presented in Section \ref{sec:conclusion}.

\section{Preintegration Theory and PI Features}
\label{sec:preintegration}
Preintegration is the method of computing motion constraint variables between keyframes in a pose graph \cite{forster2017onmanifold}. It is based on the mathematical model of the IMU and compresses the samples between frames into 9D vectors constraining the orientations, velocities, and positions of adjacent nodes in the graph. In this section, the relevant formulations for computing Forster and accurate PI features have been presented. 
\subsection{IMU Model}
The measurement model of an IMU may be expressed as follows \cite{forster2017onmanifold}:
\begin{equation}
\begin{aligned}
{ }_{\mathrm{B}}\tilde{\boldsymbol{\omega}}(t) &={ }_{\mathrm{B}} \boldsymbol{\omega}_{\mathrm{WB}}(t)+\mathbf{b}^{g}(t)+\boldsymbol{\eta}^{g}(t) \\
{ }_{\mathrm{B}}\tilde{\mathbf{a}}(t)&=\mathrm{R}_{\mathrm{wB}}^{\top}(t)\left(_{\mathrm{w}} \mathbf{a}(t)-{ }_{\mathrm{w}} \mathbf{g}\right)+\mathbf{b}^{a}(t)+\boldsymbol{\eta}^{a}(t)
\end{aligned}
\label{eq:sensor-model}
\end{equation}
where $\mathrm{R}_{\mathrm{wB}}$ represents that orientation of the IMU body frame $B$ with respect to world frame $W$, $_{\mathrm{w}} \mathbf{a}(t)$ represents the acceleration of the IMU with respect to the world frame, and $_{\mathrm{w}} \boldsymbol{\omega}_{WB}(t)$ represents the angular velocity of the IMU expressed in its local body coordinate frame. The IMU measurements ${ }_{\mathrm{B}}\tilde{\boldsymbol{\omega}}(t)$, and ${ }_{\mathrm{B}}\tilde{\mathbf{a}}(t)$ are contaminated with additive Gaussian noise $\boldsymbol{\eta}^{g}(t)$, $\boldsymbol{\eta}^{a}(t)$, and random walk biases $\mathbf{b}^{g}(t)$ and $\mathbf{b}^{g}(t)$. Furthermore, $\mathbf{g}$ in the above equation represents the known gravity direction in the world coordinate frame. 

Based on the kinematic equations of the IMU, the motion propagation of the sensor in the world frame may be expressed as follows\cite{forster2017onmanifold}:
\begin{equation}
\begin{aligned}
&\dot{\mathrm{R}}_{\mathrm{WB}}=\mathrm{R}_{\mathrm{WB}}\times  { }_{\mathrm{B}}\boldsymbol{\omega}_{\mathrm{WB}}^{\wedge}, \quad { }_{\mathrm{w}}\dot{\mathbf{v}}={ }_{\mathrm{w}} \mathbf{a}, \quad{ }_{\mathrm{w}} \dot{\mathbf{p}}={ }_{\mathrm{w}} \mathbf{v}\\ 
\end{aligned}
\label{eq:motion-model}
\end{equation}
where $(.)^\wedge$ operator converts a 3D vector into its skew symmetric matrix representation. Through integration we have: 
$$
\begin{array}{l}
\mathrm{R}_{\mathrm{wB}}(t+\Delta t)=\mathrm{R}_{\mathrm{wB}}(t) \operatorname{exp}\left(\int_{t}^{t+\Delta t}{ }_{\mathrm{B}} \boldsymbol{\omega}_{\mathrm{WB}}(\tau) d \tau\right) \\
{ }_{\mathrm{w}} \mathbf{v}(t+\Delta t)={ }_{\mathrm{w}}\mathbf{v}(t)+\int_{t}^{t+\Delta t} { }_{\mathrm{w}}\mathbf{a}(\tau) d \tau \\
{ }_{\mathrm{w}} \mathbf{p}(t+\Delta t)={ }_{\mathrm{w}}\mathbf{p}(t)+\int_{t}^{t+\Delta t} { }_{\mathrm{w}}\mathbf{v}(\tau) d \tau+\iint_{t}^{t+\Delta t} { }_{\mathrm{w}}\mathbf{a}(\tau) d \tau^{2}
\end{array}
$$
where $\Delta T$ is the sampling interval of the sensor and $\operatorname{exp}(.)$ is the $SO(3)$ exponential map.

\subsection{Forster Formulation}
Assuming the angular velocity in body frame and acceleration in world frame as constants between two IMU samples, equations \ref{eq:sensor-model} and \ref{eq:motion-model} lead to the following discrete motion propagation model:
\begin{equation}
\begin{aligned}
\mathrm{R}(t+\Delta t) &=\mathrm{R}(t) \operatorname{Exp}\left(\left(\tilde{\boldsymbol{\omega}}(t)-\mathbf{b}^{g}(t)-\boldsymbol{\eta}^{g d}(t)\right) \Delta t\right) \\
\mathbf{v}(t+\Delta t) &=\mathbf{v}(t)\\
&+\mathbf{g} \Delta t+\mathrm{R}(t)\left(\tilde{\mathbf{a}}(t)-\mathbf{b}^{a}(t)-\boldsymbol{\eta}^{a d}(t)\right) \Delta t \\
\mathbf{p}(t+\Delta t) &=\mathbf{p}(t)+\mathbf{v}(t) \Delta t+\frac{1}{2} \mathbf{g} \Delta t^{2} \\
&+\frac{1}{2} \mathrm{R}(t)\left(\tilde{\mathbf{a}}(t)-\mathbf{b}^{a}(t)-\boldsymbol{\eta}^{a d}(t)\right) \Delta t^{2}
\end{aligned}
\label{eq:forster-integration}
\end{equation}
Based on the above model, the motion corresponding to a batch of IMU samples between time steps $i$ and $j$ may be formulated as follows:
\begin{equation}
\begin{aligned}
\mathrm{R}_{j} &=\mathrm{R}_{i} \prod_{k=i}^{j-1} \operatorname{Exp}\left(\left(\tilde{\boldsymbol{\omega}}_{k}-\mathbf{b}_{k}^{g}-\boldsymbol{\eta}_{k}^{g d}\right) \Delta t\right) \\
\mathbf{v}_{j} &=\mathbf{v}_{i}+\mathbf{g} \Delta t_{i j}+\sum_{k=i}^{j-1} \mathrm{R}_{k}\left(\tilde{\mathbf{a}}_{k}-\mathbf{b}_{k}^{a}-\boldsymbol{\eta}_{k}^{a d}\right) \Delta t \\
\mathbf{p}_{j} &=\mathbf{p}_{i}\\
&+\sum_{k=i}^{j-1}\left[\mathbf{v}_{k} \Delta t+\frac{1}{2} \mathbf{g} \Delta t^{2}+\frac{1}{2} \mathrm{R}_{k}\left(\tilde{\mathbf{a}}_{k}-\mathbf{b}_{k}^{a}-\boldsymbol{\eta}_{k}^{a d}\right) \Delta t^{2}\right]
\end{aligned}
\end{equation}
As proposed in \cite{forster2017onmanifold}, through the multiplication of both sides of the above questions by $R_i$ the initial state-dependent and sensor-dependent integrated measurements can be separated:
\begin{equation}
\begin{aligned}
\Delta \mathrm{R}_{i j} & \doteq \mathrm{R}_{i}^{\top} \mathrm{R}_{j}=\prod_{k=i}^{j-1} \operatorname{exp}\left(\left(\tilde{\boldsymbol{\omega}}_{k}-\mathbf{b}_{k}^{g}-\boldsymbol{\eta}_{k}^{g d}\right) \Delta t\right) \\
\Delta \mathbf{v}_{i j} & \doteq \mathrm{R}_{i}^{\top}\left(\mathbf{v}_{j}-\mathbf{v}_{i}-\mathrm{g} \Delta t_{i j}\right)\\
&=\sum_{k=i}^{j-1} \Delta \mathrm{R}_{i k}\left(\tilde{\mathbf{a}}_{k}-\mathbf{b}_{k}^{a}-\boldsymbol{\eta}_{k}^{a d}\right) \Delta t \\
\Delta \mathbf{p}_{i j} & \doteq \mathrm{R}_{i}^{\top}\left(\mathbf{p}_{j}-\mathbf{p}_{i}-\mathbf{v}_{i} \Delta t_{i j}-\frac{1}{2} \sum_{k=i}^{j-1} \mathbf{g} \Delta t^{2}\right) \\
&=\sum_{k=i}^{j-1}\left[\Delta \mathbf{v}_{i k} \Delta t+\frac{1}{2} \Delta \mathrm{R}_{i k}\left(\tilde{\mathbf{a}}_{k}-\mathbf{b}_{k}^{a}-\boldsymbol{\eta}_{k}^{a d}\right) \Delta t^{2}\right]
\end{aligned}
\label{eq:forster-preintegration}
\end{equation}
The right hand side of the above equations represent motion constraints that are only functions of the IMU measurements. In \cite{khorrambakht2020preintegrated}, we proposed adopting these constraints as an intermediate representation for deep inertial odometry, which we called, PI features. It is worth noting that we substitute the bias-corrected values in the above equations with their corresponding raw measurements and rely on our model to compensate for their impact.
\subsection{Accurate Preintegration}
The constant world acceleration between consecutive IMU samples can be violated in the case of highly dynamic motions, or when the IMU sampling frequency is not very high. To account for this problem, \cite{henawy2019accurate} exploits the switched liner systems theory to model the state transition between each pair of the IMU samples. In other words, \cite{henawy2019accurate} assumes the acceleration and angular velocity between two IMU samples to be constant in the body frame as opposed to the world frame which is a more realistic assumption. With this assumption and after deriving the closed form solutions of the state transition between each pair of IMU samples, the following accurate preintegrated constraints are derived:
\begin{equation}
\begin{aligned}
\Delta \mathrm{R}_{i j} &=\prod_{k=i}^{j-1} \operatorname{exp}(\boldsymbol{\Theta}(t)) \\
\Delta \mathbf{v}_{i j} &=\sum_{k=i}^{j-1} \Delta \mathrm{R}_{i k} \Gamma(\boldsymbol{\Theta}(t))\left(\tilde{\mathbf{a}}_{k}-\mathbf{b}_{k}^{a}-\boldsymbol{\eta}_{k}^{a d}\right) \Delta t \\
\Delta \mathbf{p}_{i j} &=\sum_{k=i}^{j-1}\left[\Delta \mathbf{v}_{i k} \Delta t+\Delta \mathrm{R}_{i k}\Lambda(\boldsymbol{\Theta}(t))\left(\tilde{\mathbf{a}}_{k}-\mathbf{b}_{k}^{a}-\boldsymbol{\eta}_{k}^{a d}\right) \Delta t^{2}\right]
\end{aligned}
\label{eq:accurate-preintegration}
\end{equation}
where $\Lambda(.)$ and $\Gamma(.)$ are corrective terms defined in \cite{henawy2019accurate}. Furthermore $\boldsymbol{\Theta}(t)$ is defined as follows:
$$
\boldsymbol{\Theta}(t)\doteq \left(\tilde{\boldsymbol{\omega}}_{k}-\mathbf{b}_{k}^{g}-\boldsymbol{\eta}_{k}^{g d}\right) \Delta t
$$
It can be shown that when $\boldsymbol{\Theta}(t)$ is small (high sampling rate or moderate dynamics), $\Lambda(.)$ converges to $0.5$ and $\Gamma(.)$ converges to $1$. In other words, when the sampling rate of the IMU is high or the motion is highly dynamic, the two preintegrated features become identical.
\section{Experiments}
\label{sec:experiments}
In this section, we use two real-world datasets to investigate the effectiveness of accurate preintegration compared to two baselines, a model trained using raw IMU data and another using preintegrated features based on Forster's formulation. 
\begin{figure}
    %  \centering
     \begin{subfigure}[b]{0.4\textwidth}
        %  \centering
         \includegraphics[width=0.9\textwidth]{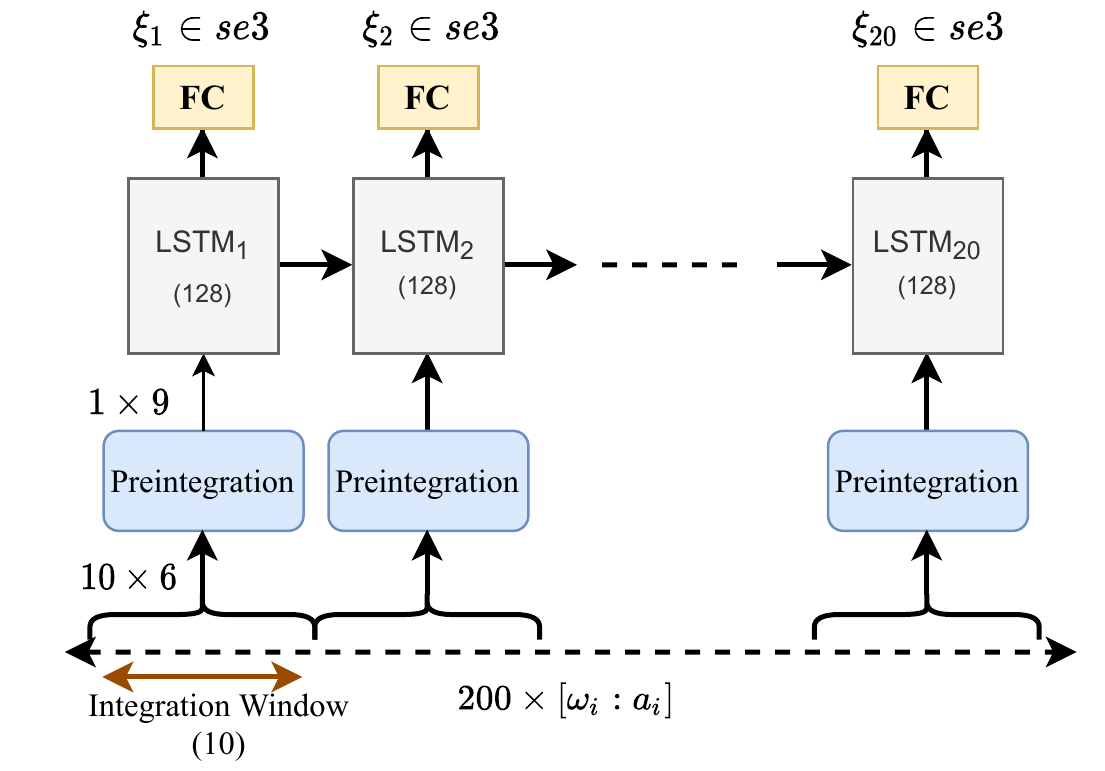}
        %  \caption{Seq. 10}
     \end{subfigure}
        \caption{The architecture of the models used in all our experiments.}
        \label{fig:architecture}
\end{figure}
\subsection{Setup}
\subsubsection{Network Architecture}
In this study, we use the base architecture from the IO-Net paper \cite{chen2018ionet}, which is a single layer bi-directional LSTM with a hidden state size of 128 and independently leaned initial hidden states. The selection of LSTM brings the flexibility of feeding inputs with various temporal resolutions without any architectural modifications. Furthermore, as reported in \cite{chen2018ionet}, bi-directionality improves the capacity of the model for capturing the motion dynamics by allowing the predictions from one step to use both past and future histories of the signal.

As depicted in Fig. \ref{fig:architecture}, we consider a temporal history of 200 IMU samples on each inference. The window length of 200 is the value for which we achieved a well balance between the performance and computational load. The presented architecture is common in all of our experiments. Two experiments are designed for each type of accurate and Forster preintegrated features. In this paper, we choose a preintegration length of 10 IMU samples based on the slowest ground-truth frequency in our datasets. Our baseline experiment bypasses the preintegration module and feeds the LSTM with the raw IMU samples. For each odometry transformation between 10 IMU samples, the hidden states of the LSTM are fed into fully connected layers with $\boldsymbol{\xi}_i \in se(3)$ predictions as outputs. These $se(3)$ vectors are converted into $SE(3)$ transformation matrices through the exponential mapping function, $\mathbf{T}_i=\exp{(\boldsymbol{\xi}_i)}$. 

\subsubsection{Training and Loss} The training loss has been formulated similar to DPC-Net\cite{valentin2018dpcnet}, with geodesic distances between the network predictions and ground-truth labels as the loss:
\begin{equation}\begin{aligned}
L&=\sum_{k=1}^{N}\boldsymbol{\Phi}_i^T\boldsymbol{\Sigma}^{-1}  \boldsymbol{\Phi}_i\\
\boldsymbol{\Phi}_i&\doteq \log(\boldsymbol{\Delta T^*}_{i,i+1}^{-1}\exp{(\boldsymbol{\xi}_i^{\wedge})})^\vee
\label{eq:loss}
\end{aligned}\end{equation}
where $\Delta\boldsymbol{T^*}_{i-1,i}=\boldsymbol{T^*}_{i-1}^{-1}\boldsymbol{T^*}_{i}$ are odometry labels, and the $\boldsymbol{\Sigma}$ is an empirical covariance matrix computed using the training data. Furthermore, $(.)^\wedge$ and $(.)^\vee$ operators respectively represent operators that transform the $se(3)$ vectors into their skew-symmetric matrix form and vise versa. 

Finally, the models are implemented using Pytorch, and Adam optimizer with an initial learning rate of $0.001$ has been adopted to train them. In order to avoid overfitting, dropout layers with a rate of $25 \%$ are added between the LSTM outputs and the FC inputs. On average, our models converged after 100 epochs. 
% We will make our implementations available to the community on the publication of the paper. 

\begin{figure}
     \centering
     \begin{subfigure}[b]{0.5\textwidth}
        %  \centering
         \includegraphics[width=1\textwidth ]{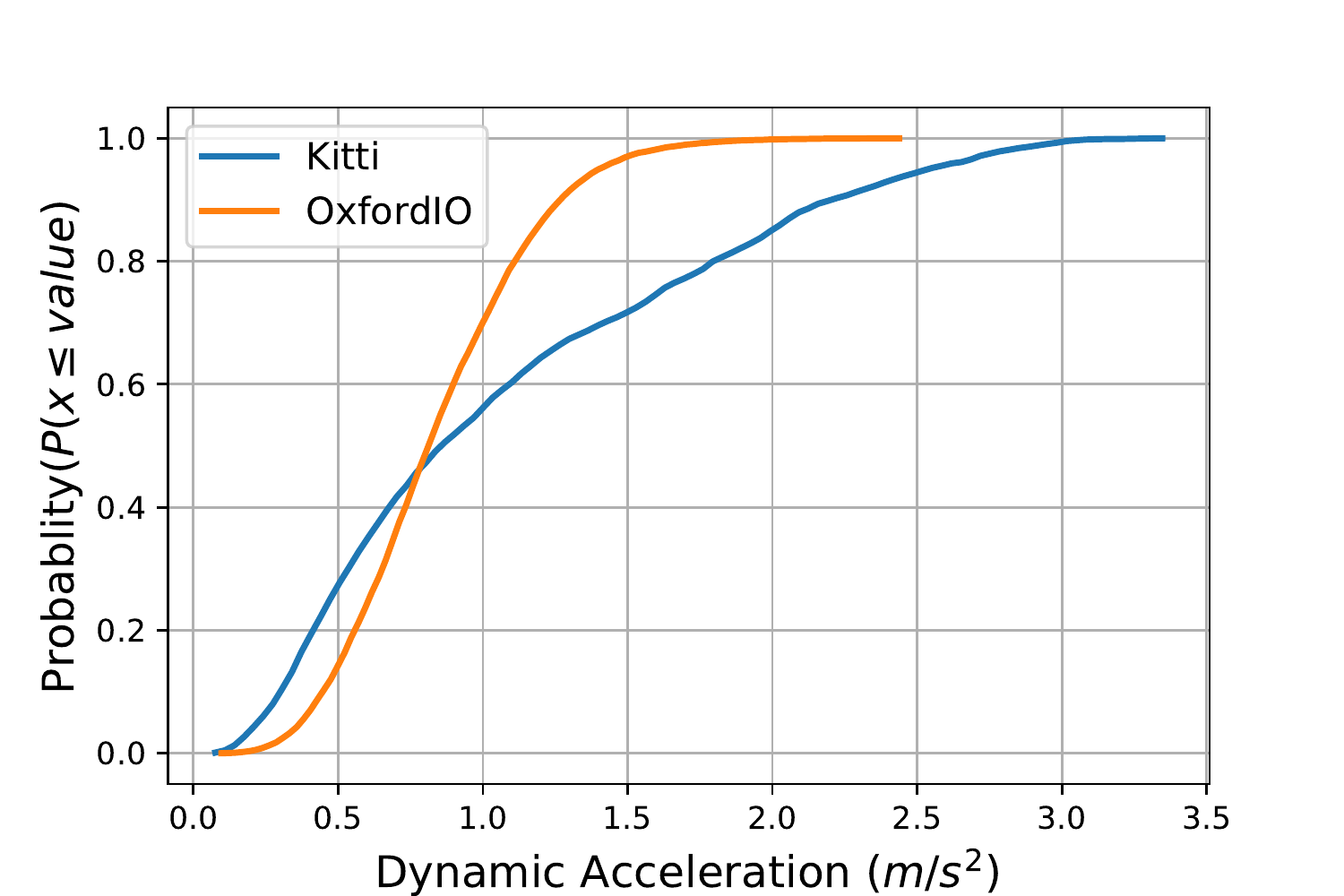}
        %  \caption{Seq. 10}
     \end{subfigure}
        \caption{The cumulative distribution of the dynamic acceleration of the Kitti and OxfordIO datasets.}
         \label{fig:dataset-dist}
\end{figure}
\subsubsection{Datasets}
We have chosen two datasets to represent fast and slow motion distributions: the OxfordIO pedestrian odometry dataset \cite{Chen2018OxIODTD} as a representative of an application domain with moderate motions, and Kitti autonomous driving dataset as an example of a domain with fast motions. Fig. \ref{fig:dataset-dist} illustrates the cumulative distribution of the static acceleration of a snippet from the test sets of the two datasets. As it can be seen in the graph, the 90th percentile of the OxfordIO dataset is at around $1.25 m/s^2$ while this value for the Kitti dataset is $2.5 ms/s^2$ with maximum accelerations of over $3 ms/s^2$, which coincides with our assumption about the fast and slow motion nature of each dataset.

In terms of sensor specifications, the Kitti dataset is recorded using a car equipped with vision, Lidar, and RTK-GPS+IMU units traversing urban and countryside environments. The IMU measurements are available at a rate of 100 Hz, and a 10 Hz centimeter-level accurate ground truth is provided through the fusion of Lidar and GPS-IMU sensors. On the other hand, the OxfordIO dataset contains the IMU readings at a rate of 100 Hz from a smartphone held in various configurations by multiple users undergoing different motion patterns. The ground-truth for this dataset is recorded using a Vicon motion capture system with millimeter-level accuracy.

\subsection{Autonomous Driving Motion Domain}
\begin{table}
\centering
\refstepcounter{table}
\caption{The performance evaluation of the three baseline models tested on Kitti dataset.}
\label{table:kitti-performance}
\begin{tabular}{c||c|c||c|c||c|c} 
\toprule
\multirow{2}{*}{\begin{tabular}[c]{@{}c@{}} test\\seq. \end{tabular}} & \multicolumn{2}{c||}{IO-Net}                                                                                               & \multicolumn{2}{c||}{\begin{tabular}[c]{@{}c@{}}PI-Net \\(Forster) \end{tabular}}                                          & \multicolumn{2}{c}{\begin{tabular}[c]{@{}c@{}}PI-Net\\(Accurate PI) \end{tabular}}                         \\
                                                                      & \begin{tabular}[c]{@{}c@{}}$t_{rel}$\\$(\%)$ \end{tabular} & \begin{tabular}[c]{@{}c@{}}$r_{rel}$\\$(deg/m)$ \end{tabular} & \begin{tabular}[c]{@{}c@{}}$t_{rel}$\\$(\%)$ \end{tabular} & \begin{tabular}[c]{@{}c@{}}$r_{rel}$\\$(deg/m)$ \end{tabular} & \begin{tabular}[c]{@{}c@{}}$t_{rel}$\\$(\%)$ \end{tabular} & \begin{tabular}[c]{@{}c@{}}$r_{rel}$\\$(deg/m)$ \end{tabular}  \\ 
\hline
10                                                                    & 11.37                                                      & 0.018                                                         & \textbf{10.23}                                             & \textbf{0.015}                                                & 10.6                                                       & 0.019                                                          \\
07                                                                    & 20.91                                                      & 0.016                                                         & 8.52                                                       & 0.013                                                         & \textbf{5.82}                                              & \textbf{0.014}                                                 \\
05                                                                    & 18.6                                                       & 0.035                                                         & 8.8                                                        & 0.021                                                         & \textbf{8.77}                                              & \textbf{0.019}                                                 \\ 
\hline
avg.                                                                  & 16.96                                                      & 0.023                                                         & 9.18                                                       & \textbf{0.0163}                                               & \textbf{8.3}                                               & 0.017                                                          \\
\bottomrule
\end{tabular}
\end{table}
This section investigates the effectiveness of accurate preintegration in IO performance improvement trained on a dynamic motion model. 
\subsubsection{Baselines} We train three models using the Kitti dataset. The base model takes the raw IMU measurements as input while the other two are fed with the two types of PI features, one computed using the accurate formulation and the other using Forster's method. Based on the 10 Hz frequency of the ground-truth labels and the 100 Hz IMU sampling rate, we set the odometry step (preintegration length) to 10 IMU samples ($100/10=10$). Furthermore, for training, sequences \{00-10\}-\{05,07,10,03\} were used and testing was performed using sequences \{05,07,10,03\}.  
\subsubsection{Evaluation Metric}
The relative translation and rotation errors defined by the KITTI benchmark \cite{Geiger2012CVPR} have been adopted as the evaluation metric. These relative errors are computed as the averaged position/orientation errors within all possible sub-sequences of lengths $100m,...,800m$. We use the open-source implementation provided in \cite{zhan2019dfvo}.  
\subsubsection{Results}
The results of this experiment have been reported in Table \ref{table:kitti-performance}. Each of the three main columns of the table represent the performance for each of the three baselines, and each row indicates the performance on each test sequence. As it can be seen, the average translation error for the model trained with accurate PI features surpasses the other two baselines, by $0.88\%$ compared to the Forster's method and $8.66\%$ compared to the model with raw input. It is important to note that models using any integration methods perform better than the baseline model with raw inputs. It is also important to note that the orientation errors of the two preintegration methods are close to each other. Because, based on Eq. \ref{eq:accurate-preintegration} and Eq. \ref{eq:forster-preintegration}, both accurate and Forster integration methods employ identical formulas to compute the rotation portion of the PI features. 
\subsection{Pedestrian Odometry Motion Domain}
Unlike the driving motion domain, pedestrians do not exhibit frequent high acceleration/deceleration and high-speed motions. Thus, in this section, we repeat the experiments on this domain to investigate the hypothesis that both accurate and Forster methods perform similarly under moderate motions. It is important to note that the sampling frequencies of the IMUs in both datasets are equal, which is important for isolating the motion characteristics as the only influential factor. 
\subsubsection{Baselines} The three baseline models of the previous section were trained on the handheld domain of the OxfordIO dataset. In order to maintain comparability, we adopted identical odometry steps and integration lengths (10 IMU samples). The sequences shown in Table \ref{tab:oxfordIO} are used for testing, and the remaining sequences were adopted for training and validation.   
\subsubsection{Evaluation Metric}
We integrate the 6-DOF $SE(3)$ odometry predictions corresponding to batches of 200 IMU samples to calculate the displacements. We compare these displacements against their ground-truth values to compute the error for each model. We then divide these errors by the displacement length to compute normalized error values.
\subsubsection{Results}
The results of this experiment have been reported in Table \ref{table:kitti-performance}. As can be seen in the table, the difference between the average performance of the two preintegration methods is marginal in this experiment. However, similar to the previous section, both preintegration methods surpass the performance of the model operating on raw data. The close gap between the performances of the two preintegration methods was expected for this motion domain. As indicated in Section \ref{sec:preintegration}, the two preintegration formulations are identical when the IMU sampling frequency is high or when the accelerations and rotation rates are not highly dynamic. 
\begin{table}
\centering
\caption{The performance evaluation of the three baseline models tested on OxfordIO dataset.}
\label{table:oxford-performance}
\label{tab:oxfordIO}
\begin{tabular}{c||c||c||c} 
\toprule
Test seq.                            & \begin{tabular}[c]{@{}c@{}}IO-Net \\~($\%$) \end{tabular} & \begin{tabular}[c]{@{}c@{}}\textbf{PI-Net }\\\textbf{(Forster)} ($\%$) \end{tabular} & \begin{tabular}[c]{@{}c@{}}\textbf{PI-Net}\\\textbf{~(Accurate PI)}\end{tabular}  \\ 
\hline
handheld-d1-s2                       & $6.5$                                                     & $\mathbf{4.94}$                                                                      & 5.0                                                                               \\
handheld-d1-s5                       & $3.33$                                                    & ${2.91}$                                                                             & \textbf{2.67}                                                                     \\
handheld-d1-s6                       & $3.12$                                                    & $\mathbf{2.82}$                                                                      & 2.91                                                                              \\
handheld-d3-s1                       & $4.48$                                                    & $\mathbf{3.42}$                                                                      & 3.6                                                                               \\
handheld-d4-s1                       & $4.32$                                                    & ${4.28}$                                                                             & \textbf{3.92}                                                                     \\
\multicolumn{1}{l||}{handheld-d4-s3} & $4.6$                                                     & $\mathbf{4.05}$                                                                      & 4.14                                                                              \\
\multicolumn{1}{l||}{handheld-d5-s1} & $3.64$                                                    & ${3.53}$                                                                             & \textbf{3.42}                                                                     \\ 
\hline
\textbf{average}                     & $4.35$                                                    & $3.71$                                                                      & $\mathbf{3.66}$                                                                              \\
\bottomrule
\end{tabular}
\end{table}

\section{Conclusion and Discussion}
\label{sec:conclusion}
Preintegration reduces the temporal dimension of the raw IMU signals by incorporating the mathematical model of the sensor. This reduction of temporal steps leads to fewer recursions by the RNN, which facilitates faster inference and better performance. 
% The integration is derived from numerical state propagation of the sensor's non-linear model, and the accuracy of this numerical solution depends on the constant world acceleration assumption between consecutive samples. However, this assumption may be violated when the motion is highly dynamic or when the sampling frequency is too low.
In this study, we investigated the impact of this numerical inaccuracy on the performance of a deep learning model trained using PI features. 
% This was carried out through two experiments on two domains with fast and slow movements.
We observed that the adoption of accurate preintegration leads to performance improvements in highly dynamic motions. In contrast, the performance gap is marginal when the movements are not highly dynamic or equivalently when the sampling frequency of the sensor is very high. 
% Nevertheless, considering the minimal additional loads of computing the accurate preintegrated features, we suggest adopting it regardless of the motion profile. 
\newpage
\bibliographystyle{IEEEtran}
\bibliography{ref.bib}
\end{document}